\documentclass[runningheads]{llncs}
\usepackage{graphicx}
\usepackage{comment}
\usepackage{amsmath,amssymb} 
\usepackage{gensymb}

\usepackage{color}
\usepackage[small]{caption}
\usepackage{url}

\usepackage[hidelinks,breaklinks]{hyperref}
\usepackage[utf8]{inputenc}
\usepackage{algorithm}
\usepackage{algorithmic}
\usepackage{epsfig}
\usepackage{subfig}
\usepackage{amstext}
\usepackage{textcomp}
\usepackage{multirow}
\usepackage{setspace}

\begin{document}
\pagestyle{headings}
\mainmatter
\def\ECCVSubNumber{2364}  

\title{Feature Lenses: Plug-and-play Neural Modules for Transformation-Invariant Visual Representations} 

\titlerunning{Feature Lenses}
%
\author{Shaohua Li\inst{1} \and
Xiuchao Sui\inst{1} \and
Jie Fu\inst{2} \and
Yong Liu\inst{1} \and
Rick Siow Mong Goh\inst{1}}
\authorrunning{S. Li et al.}
%
\institute{Institute of High Performance Computing, A*STAR, Singapore\\
\email{\{li\_shaohua,sui\_xiuchao,liuyong,gohsm\}@ihpc.a-star.edu.sg}\\
\and
Mila - Quebec Artificial Intelligence Institute, Canada\\
\email{jie.fu@polymtl.ca}}
\maketitle

\begin{abstract}
Convolutional Neural Networks (CNNs) are known to be brittle under various image transformations, including rotations, scalings, and changes of lighting conditions. We observe that the features of a transformed image are drastically different from the ones of the original image. 
To make CNNs more invariant to transformations, we propose ``Feature Lenses'', a set of ad-hoc modules that can be easily plugged into a trained model (referred to as the \textit{host model}). Each individual lens reconstructs the original features given the features of a transformed image under a particular transformation. 
These lenses jointly counteract feature distortions caused by various transformations, thus making the host model more robust without retraining. By only updating lenses, the host model is freed from iterative updating when facing new transformations absent in the training data; as feature semantics are preserved, downstream applications, such as classifiers and detectors, automatically gain robustness without retraining. Lenses are trained in a self-supervised fashion with no annotations, by minimizing a novel ``Top-$K$ Activation Contrast Loss'' between lens-transformed features and original features. Evaluated on ImageNet, MNIST-rot, and CIFAR-10, Feature Lenses show clear advantages over baseline methods.
\end{abstract}

\section{Introduction}
It is well-known that a variety of image transformations (\textit{e.g.} rotations, scalings, and changes of lighting conditions) can render Convolutional Neural Networks (CNNs) vulnerable. 
For example, our early experiments show that a pre-trained ResNet-101 model on ImageNet experiences a 20-40\% drop of classification accuracy after the image is rotated or scaled down (Table~\ref{accuracy-corr}). 

A common practice of handling image transformations is to apply these transformations to input images during training, i.e., performing data augmentation \cite{mixmatch}. Although data augmentation is effective at reducing performance deterioration, new scenarios and unseen types of transformations may be encountered occasionally, and thus the model needs to be re-trained periodically to keep its optimal performance. 
Clearly, this process is resource-inefficient and time-consuming. In addition, if too many types of transformations are applied in data augmentation, the model may perform slightly worse on original untransformed images (Section \ref{experiments}). 
More importantly, a re-trained backbone CNN may lose the original feature semantics due to parameter changes, and hence all downstream applications on top of the backbone have to be re-trained as well.

Another common practice is to do input image normalization. This works well in some scenarios. However, if the downstream applications are object detection or tracking, where multiple objects are in different scales and orientations in the same image, doing inverse rotations and scalings at the input level \cite{rcnn} requires multiple passes of the CNN feature extraction, which significantly slows down the processing. It is thus more efficient to do ``feature normalization'' at high-level convolutional layers instead.

\begin{figure*}[t]
\centering
\includegraphics[scale=0.35, viewport=-20bp 30bp 840bp 540bp, clip]{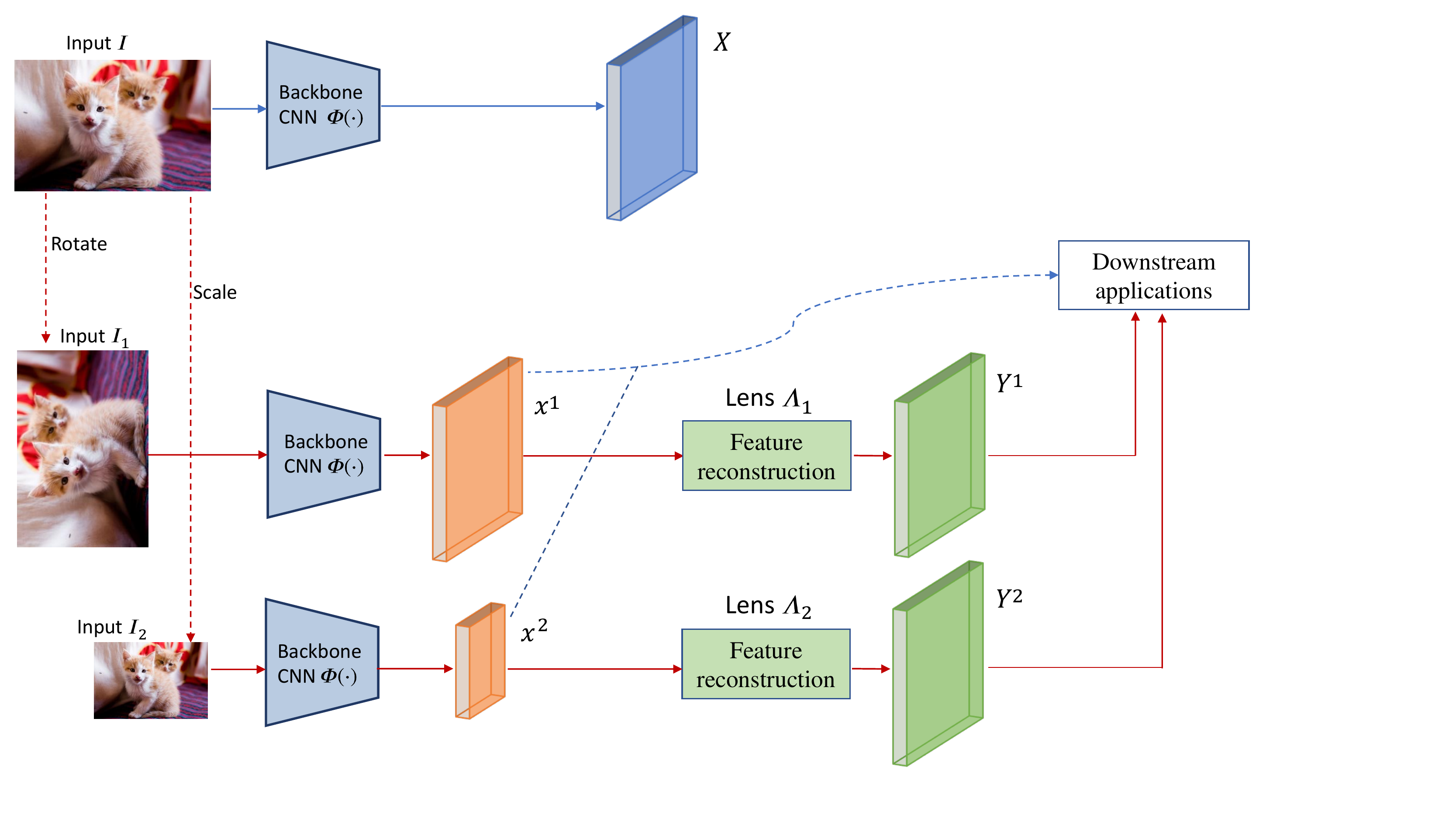}
\captionof{figure}{The blue dashed lines from the orange feature maps indicate the original pipeline. Two feature lenses $\Lambda_1$ and $\Lambda_2$ intercept features $x^1$ and $x^2$ distorted by two different input transformations, respectively, and try to reconstruct the original features $X$. The resulting features $Y^1$ and $Y^2$ are fed to the downstream applications in place of $x^1$ and $x^2$ for improved performance. Each lens module $\Lambda_i$ is a lightweight CNN that adds little overhead to the pipeline}
\label{use-flow}
\end{figure*}

To understand why transformations cause the CNN to deteriorate, we computed the correlations between the highest-layer convolutional features before and after transformations (Section~\ref{sec:corr-acc}). The feature correlations dropped significantly after transformations (Table~\ref{accuracy-corr}). Furthermore, the classification accuracy was observed to be approximately linear with the feature correlations (Fig.~\ref{corr}). These observations suggest that, if we recover the correlations between the distorted features and the original features, the CNN performance will probably be recovered accordingly, i.e., it becomes more robust against transformations.

Motivated by the analysis above, we propose a framework named ``Feature Lenses''  (Fig.~\ref{use-flow}), which consists of a set of ad-hoc modules to be plugged into a trained model (referred to as the \textit{host model} $\Phi(\cdot)$). Each individual lens $\Lambda_t$ is a lightweight CNN that tries to reconstruct the original convolutional features $X$, given the features $x^t$ of an image transformed by a transformation $t$. The reconstructed features $Y$, although not perfectly identical to $X$, are better correlated with it, and thus likely to boost downstream performance. Multiple lenses jointly counteract feature distortions caused by various transformations, thus making the host model more robust without retraining. When the host model is deployed in new scenarios or faced with new transformations, corresponding new lenses could be trained on-demand and inserted into the host model, helping the model generalize onto these new scenarios or transformations.

Feature lenses are trained in a self-supervised fashion without referring to the ground truth annotations. 
First, the original image $I$ and its transformed version $I_t$ are processed by the host model $\Phi(\cdot)$, yielding the original features $X$ and the distorted features $x^t$, respectively. Second, a lens $\Lambda_t$ transforms $x^t$ to new features $Y$. The training objective is to minimize the differences between $X$ and $Y$, measured by a feature contrast loss function.
However, when computing the feature differences, an important intuition is that the lower activations are highly noisy, and it is difficult to extract reliable information from them to guide lens training. Unfortunately, they are in overwhelming numbers and dominate the commonly used mean squared error (MSE) or mean absolute error (MAE) loss. To avoid the optimization being misled by noisy activations, we propose a novel ``Top-$K$ Activation Contrast Loss'' (TAC loss), which focuses on the most prominent activations in $X$ and $Y$, and ignores lower activations. The TAC loss shows much better empirical performance, as to be shown in our ablation studies.

To see how Feature Lenses improve the robustness of the host model, we evaluated it along with a few baseline methods on three image classification tasks, including ImageNet, MNIST-rot, and CIFAR-10. The images were transformed by one of predefined five types of transformations, including three rotations and two scalings. Feature Lenses significantly boosted the performance of the host model, outperforming other baseline methods, and approaching the accuracy of training with data-augmentation.

Feature Lenses have three practical advantages: 1) In real-world applications, from time to time, the application may receive out-of-domain images distorted by new types of transformations. By only updating lenses, the host model is freed from iterative updating. 2) As the host model has all parameters frozen, convolutional features extracted from the model maintain their original semantics. Thus downstream applications on top of the host model, such as classifiers and detectors, automatically gain robustness without being aware of the existence of lenses. 3) Lightweight lenses operate at the highest convolutional layer in the host model, and hence only adds imperceptible overhead to the whole pipeline.

To summarize, our main contributions are: 1) an on-demand plug-and-play framework named ``Feature Lenses'' to counteract various image transformations towards invariant image representations; 2) two lightweight and effective designs for rotation and scaling lenses, respectively, which can be easily extended to other transformations; 3) a self-supervised method for lens training, in which a key component is a novel ``Top-$K$ Activation Contrast Loss''. To facilitate further research, our source code is released at \url{https://github.com/askerlee/featlens}.

\section{Related Work}
A common practice is to use bigger models and train with various data augmentations, which guides CNNs to learn transformation-invariant features \cite{krizhevsky2012imagenet,mixmatch}. However, when facing new transformations, this inevitably requires re-training. 

There is an active line of works focusing on designing new layers to learn transformation-invariant features \cite{xu2014scale}. 
\cite{measure-equ} proposes to insert a feature mapping layer which aims to transform the feature maps of a rotated image back to the original feature maps, by minimizing the classification error with the same CNN weights. It resembles Feature Lenses in spirit, but the task-oriented objective is supervised.
Group invariant feature transform \cite{gift} extracts visual features that are both discriminative and robust to geometric transformations. However, this requires to extract features from multiple copies of the same input images (applied with different transformations) at test time, which is highly time-consuming.
\cite{group-equ} adopts carefully designed combinations of weight-tied convolutional kernels to preserve rotation-equivariance of features. However, one severe limitation is that the number of feature channels grows quickly as the number of layers grows, and thus the network is constrained to have just a few layers and can only handle small input images. 
\cite{roto-trans} is a slimmed-down variant of \cite{group-equ}. 
It places a roto-translation group convolution layer at the bottom of the CNN, which extracts features of various orientations. After that, the feature maps go through a few ordinary convolutional layers before being pooled into a rotation-invariant feature map for downstream tasks. As the roto-translation group convolution only locates at the bottom of the CNN, the rotation-equivariance of features only holds within small neighborhoods, and is not valid on the whole image. \cite{sfcnn} is a similar method, and is also rotation-equivariant only on small patches.



CMC \cite{contrast-multiview}, FixMatch \cite{fixmatch} and SimCLR \cite{simclr} represent a line of self-supervised contrastive learning methods for learning visual features. The training is done by requiring that the features of different views (the same image under different transformations) should agree with each other. Here transformation invariance is only a tool to force the CNN to learn high-level patterns. AVT \cite{avt} and RotNet \cite{pred-rot} do self-supervised representation learning with a different training objective: predict the transformations between two input views. Similar techniques are used in \cite{self-few-shot} to boost few-shot learning.

Feature distillation \cite{feat-distil,fitnet,detection-distil} is a variant of knowledge distillation \cite{distillation} that uses hidden features of a teacher model to train a student model. Although the training method bears some similarities with the lens training, the student model is for totally different purposes. Defensive distillation \cite{defensive-distil} distills on adversarial samples to improve model robustness against adversarial attacks. It shares similar ideas as feature lenses, although technical details are quite different.

Representation learning in the face of new transformations can be framed as a task of continual learning for domain adaptation \cite{lwf,encoder-ll,distil-retrospect,rebalancing}. Residual adapters \cite{res-adapter,adapt-count} and DAM \cite{DAM} use extra modules attached to a frozen backbone CNN to adapt to new domains. However, the controller modules totally alter the original representations, and are still trained with supervision. DATN \cite{datn} employs both supervised learning and unsupervised feature matching to adapt to new domains, but the main contributor to performance gain is still supervised learning. Backward-compatible training \cite{back-compatible} is a supervised training scheme to achieve feature compatibility between old and new models.

\section{Analysis of How Transformations Impact CNNs}
\subsection{Feature Correlations vs. Classification Accuracy}
\label{sec:corr-acc}
\begin{table*}[t]
\begin{centering}
\caption{The classification accuracy and Pearson correlations of features averaged on all ImageNet validation images, before and after transformations}
\begin{tabular}{|c|c|c|c|c|c|c|}
\hline 
 & Orig. & Rot. 90\degree & Rot. 180\degree & Rot. 270\degree & Scale 1/2 & Scale 1/3\tabularnewline
\hline 
\hline 
Top-1 Accuracy & 0.767 & 0.511 & 0.511 & 0.509 & 0.611 & 0.370\tabularnewline
\hline 
Top-5 Accuracy & 0.931 & 0.736 & 0.751 & 0.743 & 0.827 & 0.609\tabularnewline
\hline 
Whole-feature Correlation & 1.000 & 0.529 & 0.500 & 0.525 & 0.575 & 0.371\tabularnewline
\hline 
Channel-mean Correlation & 1.000 & 0.711 & 0.723 & 0.710 & 0.753 & 0.550\tabularnewline
\hline 
\end{tabular}
\label{accuracy-corr}
\par\end{centering}
\end{table*}

To gain insights into how transformations cause the CNN to deteriorate, we applied three rotations and two scalings to all the 50,000 images in the ImageNet validation set. The rotations were 90\degree, 180\degree~and 270\degree ~rotations, and the two scalings had scale factors 1/2 and 1/3, respectively.

We used a pre-trained ResNet-101 to classify these transformed images, and obtained the Top-1/Top-5 classification accuracy. In addition, to find the connections between the performance drops and the feature distortions, we used the same network to extract two types of features from the highest convolutional layer: 1. the whole $2048 \times 7\times 7$ feature maps, which are referred to as the \textbf{whole-features}; 2. as the ResNet uses global average pooled features to do classification, we also pooled the whole-features and got $2048$-dimensional \textbf{channel-mean features}. On each type of features, respectively, we computed the Pearson correlations between the features of the transformed images and the original images. Note that before computing the correlations, we first apply the dual transformations\footnote{Clockwise rotations vs. counterclockwise rotations, and upscalings vs. downscalings.}  to the feature maps to make sure the receptive field of each $(x,y)$ location in the feature maps aligns with that of the original feature maps. These classification accuracy and correlation values are shown in Table~\ref{accuracy-corr}.

\begin{figure}[th]
\centering
\subfloat[]{
{\includegraphics[scale=0.4,viewport=30 0 450 300]{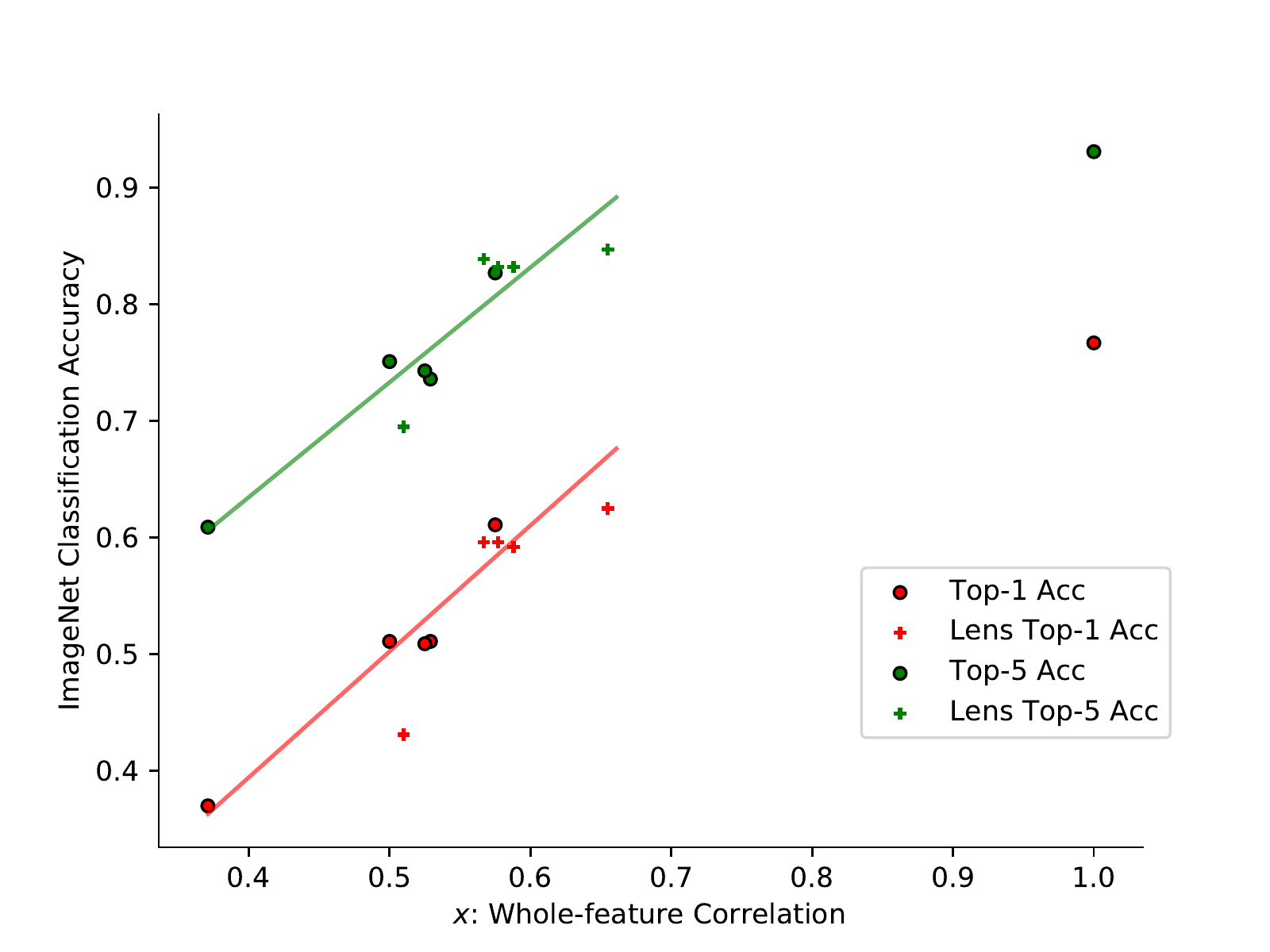} }}%
\subfloat[]{{\includegraphics[scale=0.4]{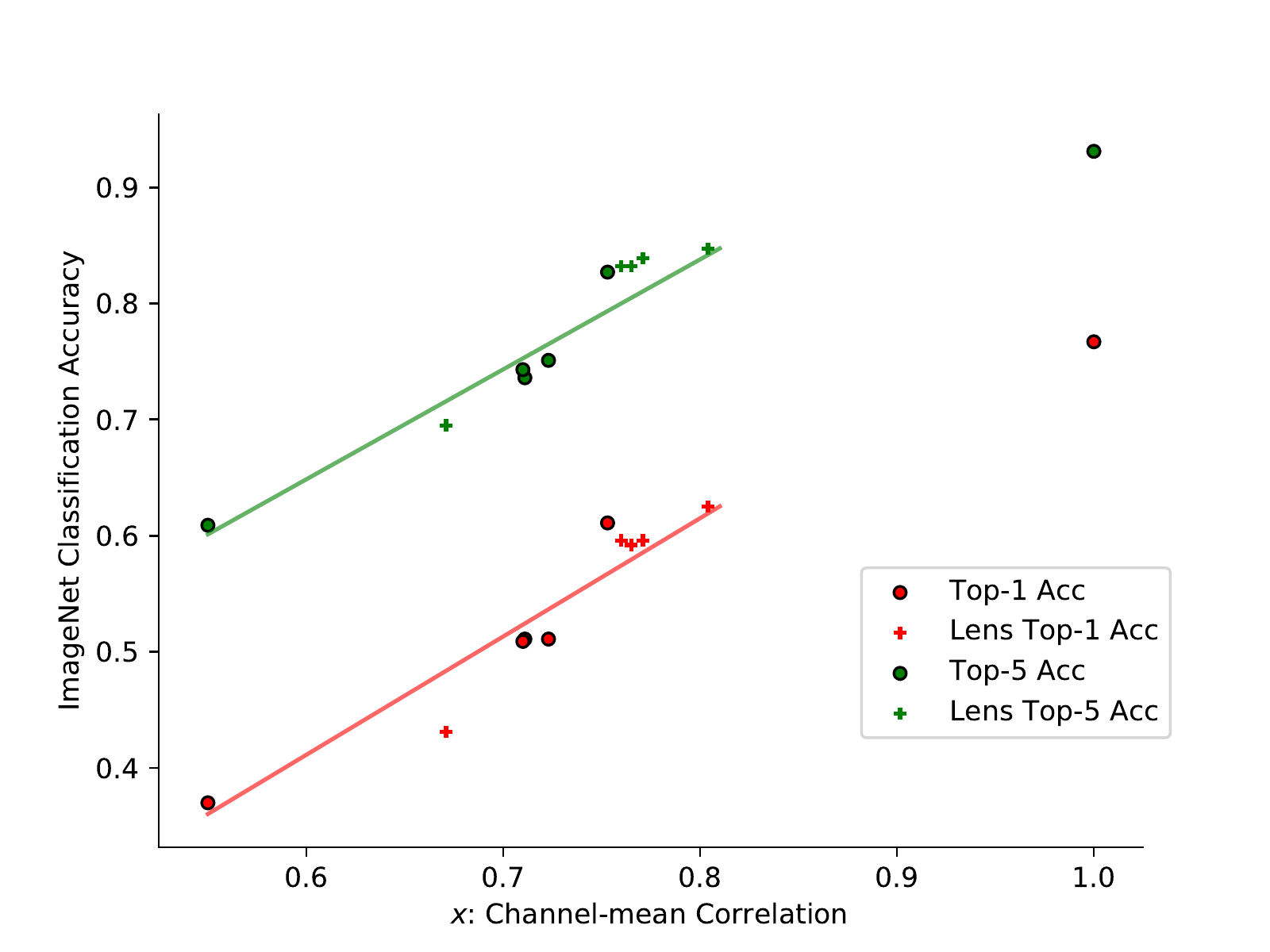} }}
\captionof{figure}{The relationships between the ImageNet classification accuracy and (a) the whole-feature correlations, or (b) the channel-mean correlations}
\label{corr}
\end{figure}

Fig.~\ref{corr} visualizes the relationships between the feature correlations and classification accuracy. If ignoring the outliers at $x=1$ (corresponding to the original images), the points in both Fig.~\ref{corr}(a) and Fig.~\ref{corr}(b) show approximately linear relationships. This observation suggests that transformations like rotations and scalings distorted the features and made them less correlated with the original images, which may be the root of the model degradation. Therefore we hypothesize that \textbf{recovering the feature correlations with original images would probably recover model performance against input transformations}.

To further validate the hypothesis above, we added the feature correlation values and classification accuracy obtained from lens-transformed features (Table~\ref{lens-scores}). These points are marked as ``\textcolor{red}{+}'' and ``\textcolor{green}{+}'' in Fig.~\ref{corr}. As can be seen, they agree well with the regressed linear relationships.

\subsection{Regularities of Feature Variations}
\vspace{-10pt}
\begin{figure}[h]
\centering
\includegraphics[scale=0.33]{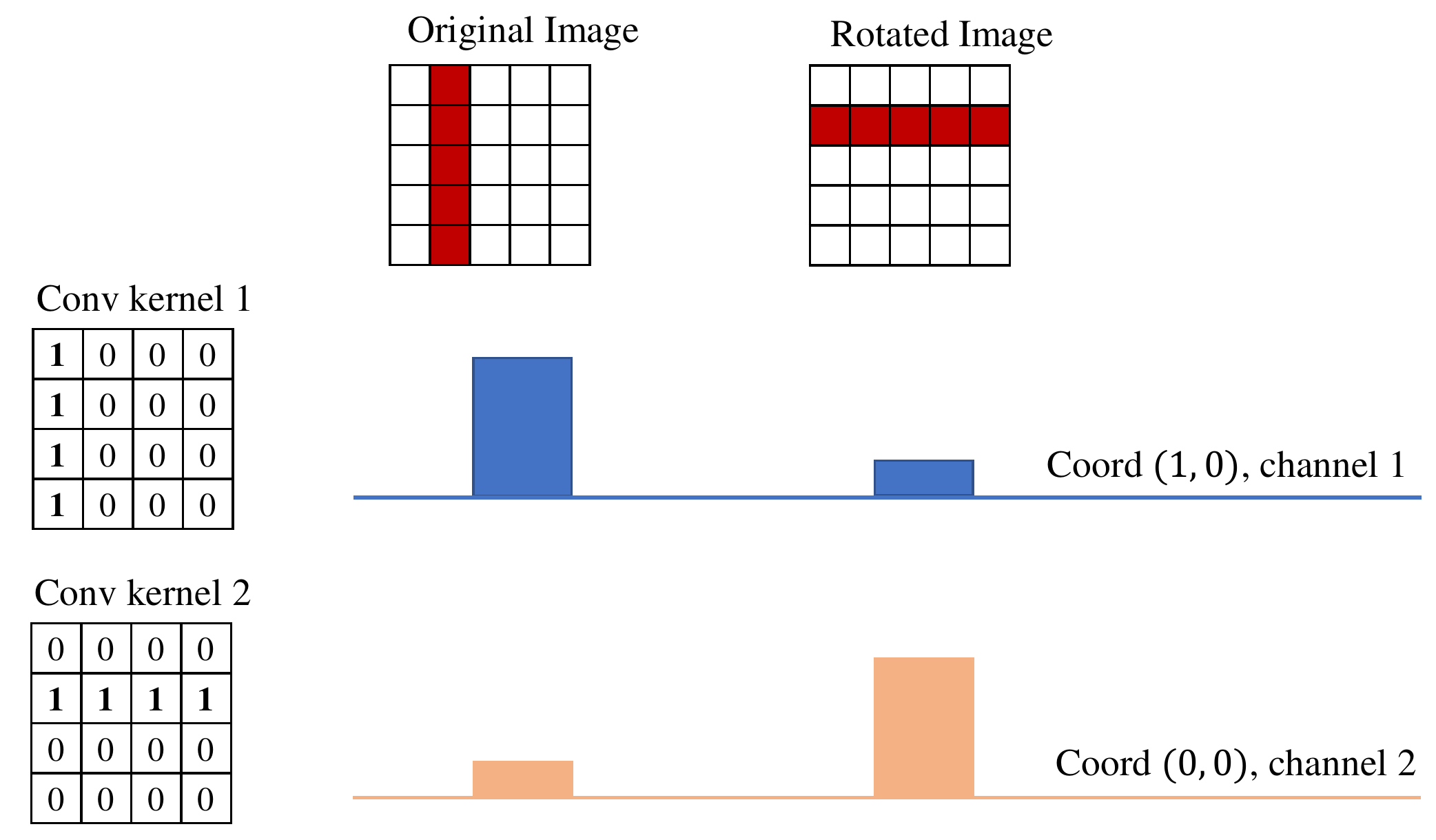}
\captionof{figure}{A rotation causes the convolutional features to change. The heights of the blue and orange bars are proportional to the feature values at the coordinate $(1,0)$ in channel 1 and $(0,0)$ in channel 2, respectively}
\label{chan-shift}
\end{figure}
\vspace{-10pt}

An interesting observation about the feature variations caused by transformations is that the variations follow regular patterns. To illustrate this point, a simple example is shown in Fig.~\ref{chan-shift}. The original image contains a vertical bar, on which the kernel 1 outputs a high feature value at coordinate $(1,0)$, and the kernel 2 outputs a low value at $(0,0)$. After rotation, it becomes a horizontal bar. 
Then the kernel 1 outputs a low feature value at $(1,0)$, and instead the kernel 2 outputs a high value at $(0,0)$. Similar feature variations happen to many tiny patterns in an image, which operate in a deterministic and predictable way. These variations conspire to cause the feature correlations to drop, as described in Section \ref{sec:corr-acc}. 

Since the feature variations are deterministic and predictable, in theory, if knowing the type of transformation, we should be able to recover the original features given varied features. 
However, as the number of convolutional layers increases and some information is discarded by ReLU operators, how the features vary becomes overly complicated and indiscernible. 
Nonetheless, we can still treat the feature variation process as a black box, and solve the inverse problem with a neural network, which is the motivation behind Feature Lenses. 

\section{Designs of Feature Lenses}
In principle, Feature lenses can be applied to many types of CNN architectures. As a proof of concept, here we present the designs for ResNet, but it would be straightforward to adapt them to other architectures. 
To minimize the overhead brought on to the host model, we adopt lightweight designs for lenses with only $1\sim 2$ convolutional layers. 
All lenses receive features as their input from the last bottleneck block (or a basic block in smaller models) in a ResNet.


\subsection{Rotation Lens}
\begin{figure*}[t]
\includegraphics[scale=0.35,viewport=20 95 890  540]{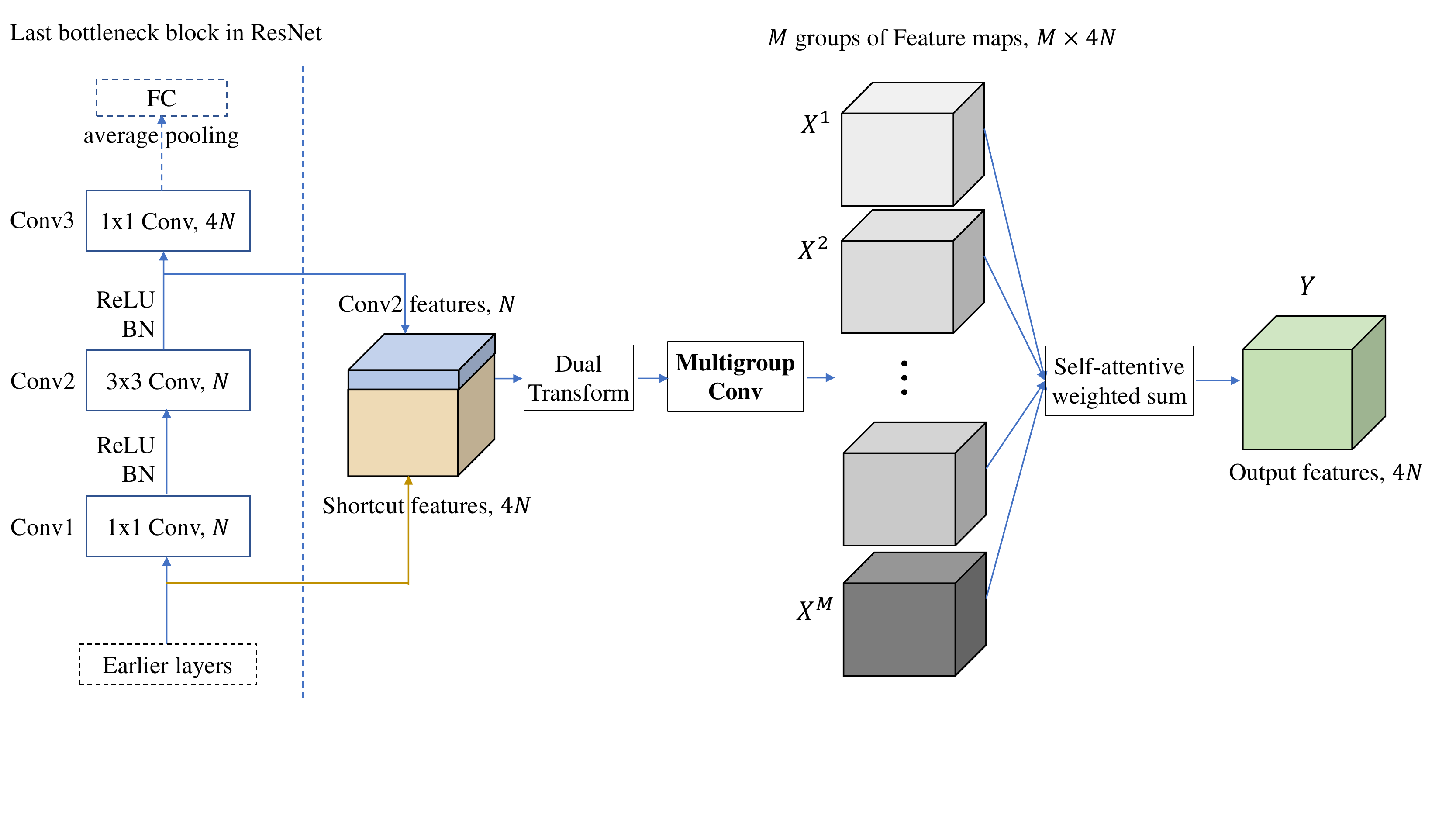}
\centering
\captionof{figure}{A rotation lens attached to the last bottleneck block of a ResNet. It consists of: 1) a dual transformation operator that rotates the feature maps back to an orientation aligned with the original feature maps, 2) a multigroup convolution layer that outputs $M$ groups of feature maps, each with $4N$ channels, and 3) a self-attentive weighted sum that adds up the $M \times 4N$ feature maps across groups to produce a $4N$-channel output feature maps. $N=512$ for the ResNet-101}
\label{rot-lens}
\end{figure*}

A rotation lens is used when the image rotation is detected, and the rotation angle falls within a predefined range. In a bottleneck block \cite{resnet}, there are three convolutional layers: Conv1, Conv2, and Conv3. A rotation lens is designed to be a surrogate module of Conv3, i.e., to replaces Conv3 of the last block. So it receives the same features as Conv3: the output features $x_2$ of Conv2 ($N$ channels), as well as the shortcut features $x_0$ from the previous bottleneck block ($3N$ channels). Here $N=512$ for ResNet-101. These two sets of features are concatenated along the channel dimension to form the input ($4N$ channels) to the lens for further transformations.

A rotation lens consists of three components: 
\begin{enumerate}\vspace{-6pt}
    \item A dual transformation operator that reverses the rotation applied to the input image, so that the orientations of the features are aligned with the original feature maps, to facilitate computing the feature contrast loss;
    \item A multigroup convolution layer that receives the rotated ResNet features, and outputs $M$ groups of feature maps, each with $4N$ channels;
    \item A self-attentive weighted sum, which aggregates the $M$ groups of features above into one group, with $4N$ channels.
\end{enumerate}\vspace{-4pt}

For the dual transformation operator, the rotation angle is the inverse angle of a predefined angle designated to this lens.

The multigroup convolution layer is an ordinary convolution layer, whose outputs are split into multiple groups of feature maps. The multiple groups are then aggregated into one group with a self-attentive weighted sum. The purpose of the multiple groups is to make the lens more expressive for modeling complex nonlinear transformations. The multigroup convolution layer has a kernel size of $1\times 1$, which does cross-channel transformations of the feature vector at each pixel in the feature maps, without considering its context. We have experimented with kernels of $3\times 3$, which did not perform as well as $1\times 1$ kernels. This suggests that the feature vector at each pixel has contained most information for recovering the original features, and incorporating its context does not help much.

The self-attentive weighted sum can be viewed as a soft version of max pooling across the $M$ groups of feature maps. Denoting the $M$ groups of feature maps as $X^1,\cdots,X^M$, their self-attentive weighted sum $Y$ is defined as:%
\begin{align} \label{attn-sum}
    X^m =&\, \text{Conv}_m([x_2, x_0]), \nonumber \\
    Y_{chw} =&\, \text{softmax}(X_{chw}^\cdot)^\intercal \cdot X_{chw}^\cdot,
\end{align} where $[x_2, x_0]$ is the concatenation of Conv2 and shortcut features in the last bottleneck block, $X_{chw}^m$ is the feature value at coordinate $(h,w)$ in the $c$-th channel of the $m$-th group, and $X_{chw}^\cdot$ is the column vector $(X_{chw}^1,\cdots, X_{chw}^M)$. 

\subsection{Scaling Lens}
\begin{figure*}[h]
\includegraphics[scale=0.35,viewport=0 95 920  540,clip]{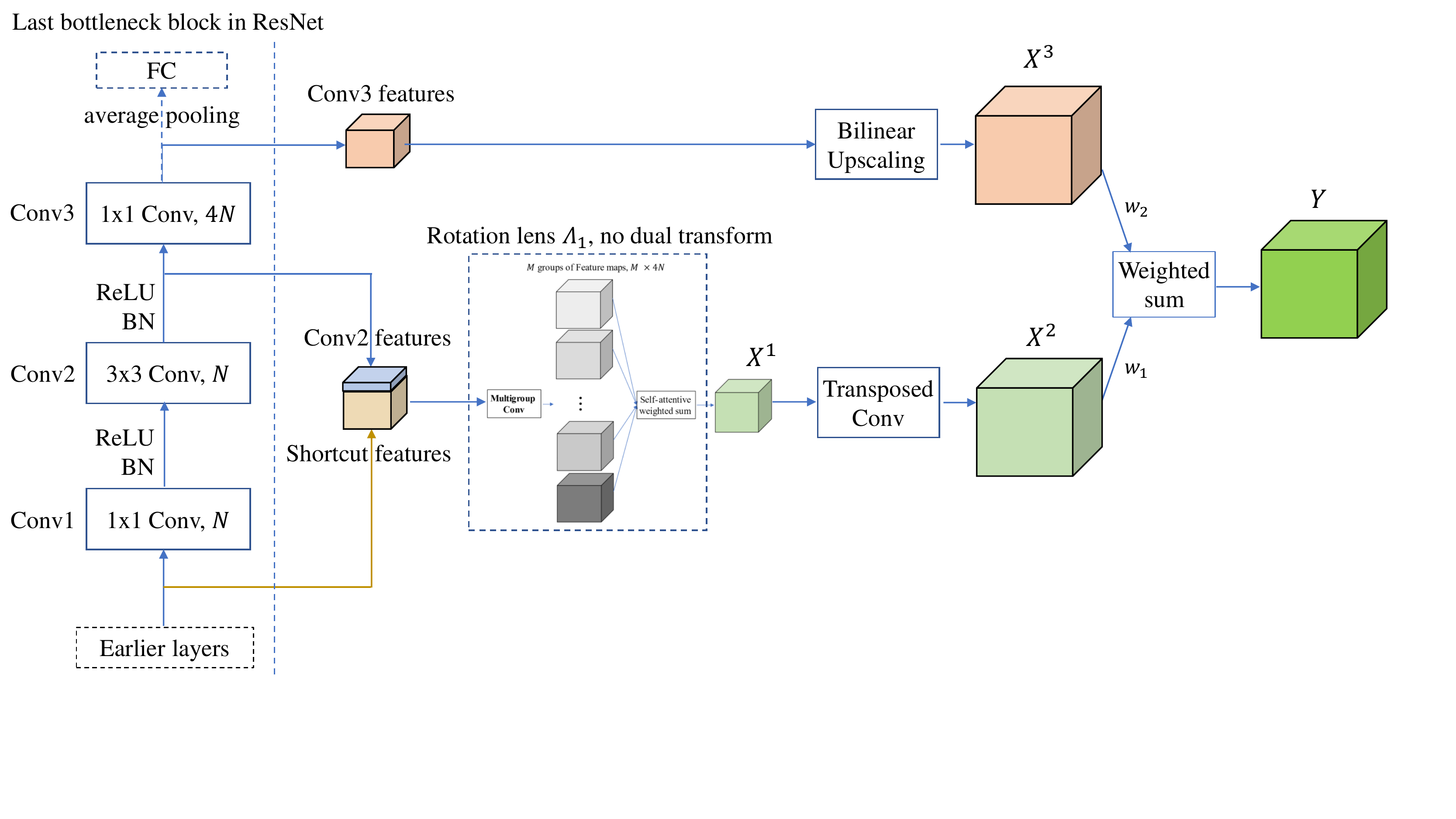}
\centering
\captionof{figure}{A scaling lens attached to the last bottleneck block of a ResNet. It consists of: 1) a modified rotation lens with the dual transformation operator removed, which does cross-channel transformations, 2) a transposed convolution layer that upsamples the feature maps $X^1$ to $X^2$ of the same size as the original feature maps, 3) a bilinear upscaling operator that upsamples the Conv3 feature maps to the same size of the original feature maps, and 4) a weighted sum that combines the two sets of features}
\label{scale-lens}
\end{figure*}
A scaling lens is used when the image is scaled, and the scale factor falls within a certain range. Here we focus on downscalings, which are more frequently encountered than upscalings. The architecture of the scaling lens is extended from the rotation lens, by adding features from Conv3 as residual features.

Same as the rotation lens, a scaling lens receives the output features $x_2$ of Conv2 in the last bottleneck block, as well as the shortcut features $x_0$ from the previous bottleneck block. These features are concatenated and fed into an embedded lens $\Lambda_1$, which is derived from a rotation lens by removing the dual transformation operator. Lens $\Lambda_1$ applies cross-channel transformations, whose output feature maps $X^1$ are upsampled to the same size as the original feature maps (from the input image without scaling) by a transposed convolution layer. In addition, the output feature maps $x_3$ from Conv3 in the last bottleneck block are upscaled with a bilinear interpolation operator, and serves as residual features. These two sets of feature maps are linearly combined:\vspace{-5pt}
\begin{align}
    X^2 =&\; \text{TransConv}(\Lambda_1([x_2, x_0])), \nonumber \\
    Y =&\; w_1 \cdot X^2 + w_2 \cdot \text{bilinear}(x_3),\vspace{-5pt}
\end{align}
where $w_1, w_2$ are two learnable positive scalar weights satisfying $w_1+w_2 = 1$.

The incorporation of the Conv3 features is based on the observation that the feature map correlations between scaled-down feature maps and original feature maps are relatively high, and combining them allows the lens focus on the nonlinearities introduced by the scaling.

\section{Top-$K$ Activation Contrast Loss for Self-Supervision}
\begin{figure*}[h]
\centering
\includegraphics[scale=0.35]{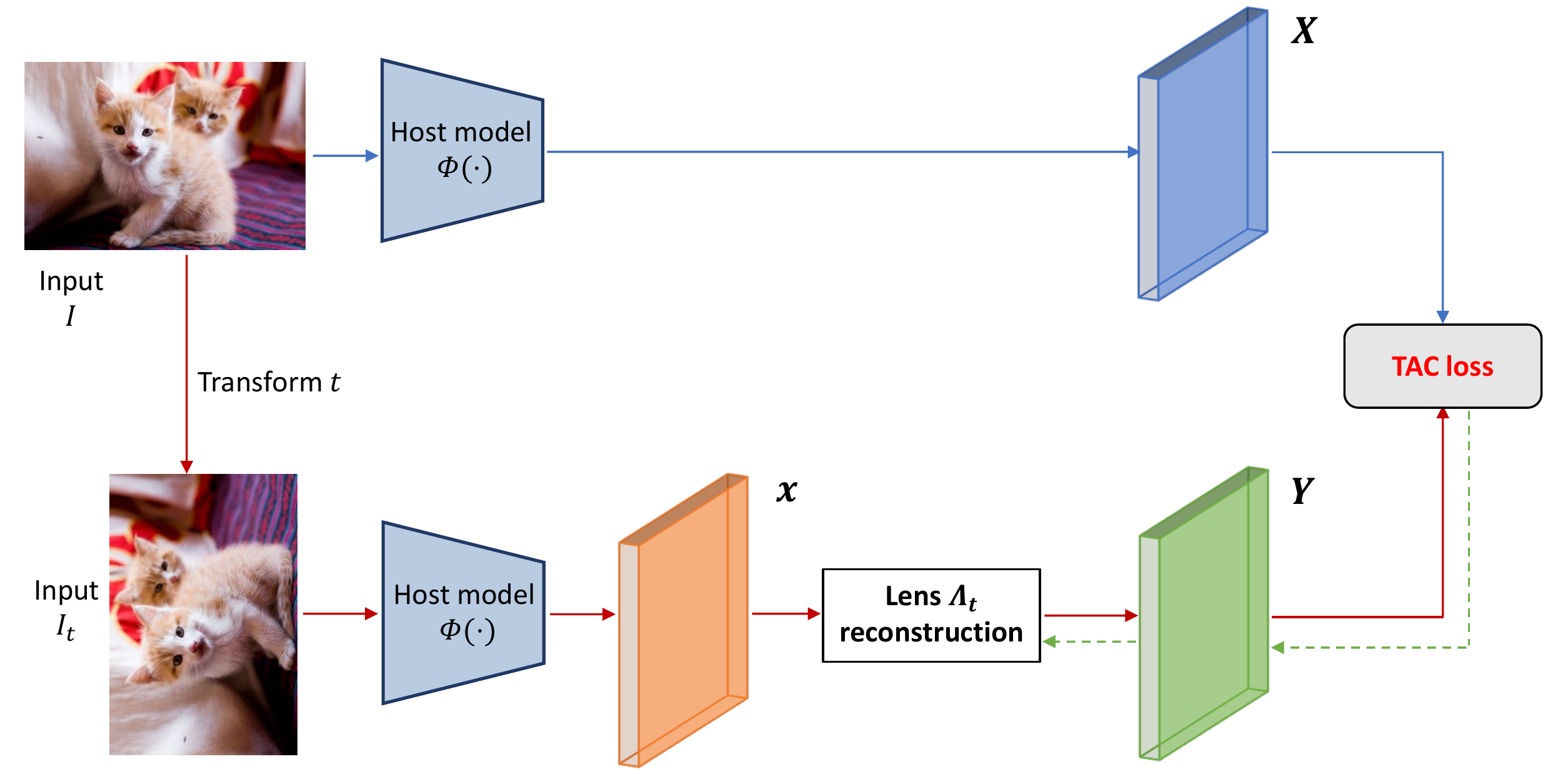}
\captionof{figure}{Self-supervised training flow of a rotation lens $\Lambda_t$. It reconstructs the distorted image features $x$ to $Y$, which are contrasted with the original image features $X$ by the Top-$K$ Activation Contrast (TAC) loss. The dashed green arrows indicate the flow of gradients. During training, all parameters in the host model are frozen}
\label{lens-sketch}
\end{figure*}

Feature lenses are trained in a self-supervised fashion without referring to the ground truth annotations. 
Firstly, the original image $I$ and its transformed version $I_t$ are processed by the host model $\Phi(\cdot)$, yielding the original features $X$ and the distorted features $x$, respectively. Secondly, a lens $\Lambda_t$ transforms $x$ to new features $Y$. The training objective is to minimize the difference between $X$ and $Y$, measured by a feature contrast loss function. All parameters in the host model are frozen, and only the lens parameters are updated.

More formally, let us denote a lens with parameters $\theta$ as $\Lambda_\theta$. Given a set of distorted training features $\{x^i\}$, the self-supervised training objective of a lens is to find the optimal $\theta$, so that $\Lambda_\theta$ minimizes the reconstruction error between the original features $\{X^i\}$ and reconstructed features $\{Y^i=\Lambda_\theta(x^i)\}$:\vspace{-3pt}\begin{equation}
\theta^* = \arg\min_\theta \textstyle\sum_i \mathcal{L}(X^i, \Lambda_\theta(x^i)).
\end{equation}\vspace{-8pt}

The mean squared error (MSE) and mean absolute error (MAE) are commonly used to measure reconstruction errors at the image pixel level. However, when applied to high-level convolutional features, they suffer from massive noises within the features. Convolutional activations are highly noisy, in that small positive activations do not guarantee that the patterns encoded by the convolutional filters exist in the image. In the example given in Fig.~\ref{chan-shift}, a horizontal bar induces positive activations of the kernel for detecting vertical bars. Accordingly, only the highest and lowest activations most reliably indicate the existence of specific interesting patterns the CNN encodes. As the majority values in a feature map are low activations (either positive or negative), these activations would dominate the MSE/MAE summation and probably mislead the optimization process towards bad local minima. Our ablation study (Section \ref{sec:ablation}) has shown that using MSE/MAE does not yield satisfactory performance. 

To direct a lens at reconstructing informative activations, we propose a novel ``Top-$K$ Activation Contrast Loss'' (TAC loss) to better measure the difference between feature maps. It focuses on the most prominent activations in $X$ and $Y$, and ignores low activations. Here the prominent activations include both the top positive and top negative activations (indicating ``negative patterns''). 

The TAC loss mainly consists of four terms: the positive overshoot and undershoot losses $\mathcal{L}^{+>}$ and $\mathcal{L}^{+<}$ (indicating that the reconstructed positive features are too big/small), and their negative counterparts $\mathcal{L}^{->}$ and $\mathcal{L}^{-<}$. On the $c$-th channel, the highest- and the lowest-$K$ activations in $X_c$ are selected at $p_1,\cdots, p_K$, and $q_1,\cdots, q_K$, respectively, where each $p_k$ or $q_k$ is a coordinate pair $(h,w)$. The four losses are defined as:\vspace{-5pt}%
\begin{align}
    \mathcal{L}^{+>}_c(X,Y) &= \textstyle\sum_{p_k} |X_{c,p_k} - Y_{c,p_k}| \quad \text{where } X_{c,p_k} < Y_{c,p_k}, \nonumber \\
    \mathcal{L}^{+<}_c(X,Y) &= \textstyle\sum_{p_k} |X_{c,p_k} - Y_{c,p_k}| \quad \text{where } X_{c,p_k} > Y_{c,p_k}, \nonumber \\
    \mathcal{L}^{->}_c(X,Y) &= \textstyle\sum_{q_k} |X_{c,q_k} - Y_{c,q_k}| \quad \text{where } X_{c,q_k} > Y_{c,q_k}, \nonumber \\
    \mathcal{L}^{-<}_c(X,Y) &= \textstyle\sum_{q_k} |X_{c,q_k} - Y_{c,q_k}| \quad \text{where } X_{c,q_k} < Y_{c,q_k}.
\end{align}\vspace{-10pt}

In addition, when there are big activations in $Y$ where the corresponding activations in $X$ are small, these activations are not selected into the top-$K$ of $X$, thus are unconstrained. Such activations should be suppressed with another loss term. It is similar to the positive overshoot loss, except that the activations are the top-$K$ in $Y$. This loss term is denoted as $\mathcal{L}^{+>}_c(Y,X)$, and is incorporated in the TAC loss as well. The TAC loss is a sum over all channels:\vspace{-5pt} 
\begin{equation}\setlength{\jot}{-5pt}
\begin{split}
    \mathcal{L}_\text{tac}(X,Y) = \sum_c \Big\{ & \mathcal{L}^{+<}_c(X,Y) + \mathcal{L}^{-<}_c(X,Y) \\ &+ d_1 \big(\mathcal{L}^{+>}_c(X,Y) + \mathcal{L}^{->}_c(X,Y) +  \mathcal{L}^{+>}_c(Y,X) \big) \Big\},
\end{split}
\end{equation}
where $d_1 < 1$ is a discount factor to balance overshoot and undershoot losses.

\section{A Few Practical Issues when Using Lenses}\vspace{-5pt}
\subsection{Granularity of Transformations for Lenses}\vspace{-5pt}
In principle, if two transformations cause different feature variations, ideally they should be handled with two different lenses. For example, 90\degree, 180\degree~and 270\degree ~rotations are handled by three different lenses. However, in practice, if we use many lenses, each accommodating a narrow range of transformations, the marginal gains decrease quickly, and the deployment cost increases.

In our experiments, we only trained three rotation lenses to handle 90\degree, 180\degree~and 270\degree~rotations, respectively. Considering that in the MNIST-rot dataset, the rotation angles of images could be any random value within [0, 360), we bin a random angle into one of the four intervals: [-45, 45), [45, 135), [135, 225) and [225, 315), and handle the image by either the original lens (which passes through the features), or one of the three rotation lenses. The classification error with predicted lenses was reasonably low ($< 14\%$, Section \ref{experiments}).

\subsection{Estimation of the Transformation Type}\vspace{-5pt}
In order to apply the most appropriate lens, one needs to know which type of transformation has been applied to the input image. Unfortunately, in practice, such information is usually unavailable, and can only be estimated by classifiers or heuristic rules.

For instance, in order to choose the rotation lenses, we used a simple FC layer on top of the lens input features to classify the rotation type. On both ImageNet and MNIST-rot, rotation classification accuracy of $70\sim 80\%$ was achieved. Although not very high, they showed to perform only slightly worse than choosing the lens by using the ground-truth rotation angle. For non-geometric transformations, such as blurring or the lighting condition changes from day to night, we could also employ classifiers to decide which transformation has been applied.

Regarding the scaling lenses, object detection and tracking are their prospective application scenarios, where the detected objects in images/videos are intrinsically in multi-scales. In a common detection pipeline, ROI-pooling \cite{faster-rcnn} interpolates the feature maps of proposed regions to a standard size, before the features are fed to downstream classifiers. As the ROI-pooling does not address feature distortions caused by scalings \cite{pedestrian-detect}, scaling lenses could serve as a drop-in replacement of ROI-pooling. In such cases, the scaling ratio can be easily computed as the ratio between the proposal size and the standard feature size.

\subsection{When Paired Images are Unavailable}\vspace{-5pt}
The lens training requires paired images before and after a particular transformation. For rotations and scalings, such pairs are readily generated. However, paired images are usually unavailable for some transformations, such as changes of lighting conditions. In this case, we could train a Cycle-Consistent Generative Adversarial Network (CycleGAN) \cite{cyclegan} on two sets of unpaired images, e.g. one set of day images and one set of night images, and use the CycleGAN to translate each day image to the paired night image \cite{input-adapter}. 

\section{Experiments}
\label{experiments}

We evaluated Feature Lenses and a few baseline models on three image classification tasks: ImageNet-2012, MNIST-rot \cite{mnist-rot} and CIFAR-10. The transformations applied to each dataset are listed as below:\vspace{-6pt}
\begin{itemize}
    \item \textbf{ImageNet}: Random 90\degree, 180\degree~and 270\degree ~rotations, and 1/2 and 1/3 scalings.
    \item \textbf{MNIST-rot}: Random rotations of angles uniformly sampled from [0, 360).
    \item \textbf{CIFAR-10}: Random 90\degree, 180\degree~and 270\degree ~rotations.
\end{itemize}\vspace{-6pt}

\subsection{Ablation Studies} \label{sec:ablation}
We carried out two ablation studies on ImageNet with random rotations, to investigate: 1) how the TAC loss performs compared with MAE/MSE, and 2) how much worse lenses perform by selecting rotation lenses by predicted rotation angles, compared with selections by the ground truth angles.

In the first ablation study, Feature Lenses were trained using MSE, MAE, TAC, and their combinations. The ``MSE+TAC'' is a combination of MSE and TAC, i.e., $0.5\cdot\text{MSE}(X,Y)+0.5\cdot\mathcal{L}_\text{tac}(X,Y)$. Similarly, ``MAE+TAC'' is a combination of MAE and TAC.

Table~\ref{ablation}(a) shows the Top-1 accuracy obtained using different losses. Using MSE or MAE alone, the model always stayed at the initial random states without outputting useful features. When combined with TAC, they performed slightly worse than using TAC alone. This result supports our intuition that low activations mislead the optimization process.

In the second ablation study, the accuracy of rotation prediction by Feature Lenses was evaluated, and shown in the upper part of Table~\ref{ablation}(b). The ImageNet accuracy under the two settings was reported in the lower part of Table~\ref{ablation}(b). When using the predicted rotation to select the rotation lens, Feature Lenses only performed 0.4\% worse than using the ground truth rotation. Considering the rotation prediction accuracy is not very high, this performance gap is somewhat surprising. One possible explanation is that the image features that are difficult for rotation classification may also be less impacted by input rotations, and thus transforming them with a wrong lens still leads to similar accuracy. 

\begin{table}[t]
\captionof{table}{Ablation performance of Feature Lenses on ImageNet}
\centering
\subfloat[MAE/MSE vs. the TAC loss]{
\begin{tabular}{c|c|c|c}
\hline 
Top-1 Acc. & Rot 90 & Rot 180 & Rot 270 \tabularnewline
\hline 
\hline 
MSE & 0.010 & 0.010 & 0.010 \tabularnewline
\hline 
MSE+TAC & 0.590 & 0.591 & 0.594 \tabularnewline
\hline 
MAE & 0.010 & 0.010 & 0.010 \tabularnewline
\hline 
MAE+TAC & 0.589 & 0.589 & 0.597 \tabularnewline
\hline 
TAC & \textbf{0.600} & \textbf{0.597} & \textbf{0.598} \tabularnewline
\hline 
\end{tabular}
}%
\subfloat[Predicted vs. true rotations]{
\begin{tabular}{c|c|c|c}
\hline 
Pred Rot Acc. & Rot 90 & Rot 180 & Rot 270\tabularnewline
\hline 
\hline 
 & 0.760 & 0.807 & 0.760\tabularnewline
\hline 
\multicolumn{1}{c}{} & \multicolumn{1}{c}{} & \multicolumn{1}{c}{} & \tabularnewline
\hline 
Top-1 Acc. & Rot 90 & Rot 180 & Rot 270\tabularnewline
\hline 
\hline 
Use Pred Rot & 0.600 & 0.597 & 0.598\tabularnewline
\hline 
Use True Rot & 0.604 & 0.601 & 0.603\tabularnewline
\hline 
\end{tabular}
}
\label{ablation}
\end{table}\vspace{-4pt}

\subsection{Comparison with Other Methods}
\subsubsection{Evaluated Methods}
The following methods were evaluated:\vspace{-6pt}
\begin{itemize}
    \item \textbf{ResNet}: The original pre-trained models provided by PyTorch.
    \item \textbf{DataAug}: ResNet Models trained on images randomly augmented by the evaluated transformations.
    \item \textbf{Xlayer}: Pre-trained ResNet expanded with an extra bottleneck/basic block on top of the last conv block. The original model weights (including FC) are frozen, and only the extra block is updated to improve classification accuracy on distorted features. It is trained with supervision from image class labels. It is an instance of the task-oriented regression method presented in \cite{measure-equ}.
    \item \textbf{RP\_RF\_1} \cite{backrot}: A rotation-invariant CNN by feature map back-rotation.
    \item \textbf{ORN} \cite{ORN}: Oriented response network that learns rotation-invariant features.
    \item \textbf{Lenses}: Feature Lenses attached to pre-trained ResNet models. 5 lenses were trained: 90\degree, 180\degree~and 270\degree ~rotations; 1/2 and 1/3 scalings. A simple FC classifier is used to predict what type of rotation the current image is applied, and  the rotation lens is selected accordingly.
\end{itemize}\vspace{-3pt}

On ImageNet, all models were based on ResNet-101. On MNIST-rot and CIFAR-10, all models were based on ResNet-18. To improve the performance of ResNet-18 on small images, we resized image resolutions to $56\times 56$ and $64\times 64$ for MNIST-rot and CIFAR-10, respectively, and removed the first pooling layer.

Due to technical challenges rooted in their own algorithms, RP\_RF\_1 and ORN were only implemented as small models and evaluated on MNIST-rot and CIFAR-10. RP\_RF\_1 and ORN were trained on the original MNIST and CIFAR-10 datasets, to evaluate how rotation-invariant the learned features were.

\subsubsection{Evaluation Results}
We evaluated the six methods on ImageNet, MNIST-rot and CIFAR-10. Table~\ref{lens-scores} presents their Top-1 classification accuracy. 

\begin{table}[t]
\begin{centering}
\caption{Evaluation results of six methods on ImageNet, MNIST-rot and CIFAR-10}
\begin{tabular}{c|c|c|c|c|c|c}
\hline 
 & \multicolumn{4}{c|}{ImageNet} & \multirow{2}{*}{MNIST-rot} & \multirow{2}{*}{CIFAR-10}\tabularnewline
\cline{1-5} 
          & Orig. & Rot   & Scale 1/2 & Scale 1/3 &  & \tabularnewline
\hline 
\hline 
ResNet    & 0.767 & 0.510 &   0.602   &  0.358    & 0.320 & 0.376 \tabularnewline
\hline 
DataAug   & 0.744 & 0.632 &   0.634   &  0.478    & 0.986 & 0.823 \tabularnewline
\hline 
Xlayer    & 0.767 & 0.592 &   0.626   &  0.410    & 0.822 & 0.531 \tabularnewline
\hline 
RP\_RF\_1 & -     & -     &     -     &   -       & 0.842 & 0.441 \tabularnewline
\hline 
ORN       & -     & -     &     -     &   -       & 0.574 & 0.407 \tabularnewline
\hline 
Lenses    & 0.767 & 0.598 &   0.625   &  0.431    & 0.866 & 0.695 \tabularnewline
\hline 
\end{tabular}
\label{lens-scores}
\par\end{centering}
\end{table}

It can be seen that on transformed images, both DataAug and Lenses have significantly improved the original models. DataAug always performed better than Lenses with a clear margin. However, on the original ImageNet, the accuracy of DataAug dropped 2\% compared with models trained without rotations or scalings. It suggests that transformed image patterns compete with original image patterns for the limited model capacity. In this aspect, Lenses show the advantage that they do not compete with the original images, but always enhance the host CNN without sacrificing the performance on common input images. 

Despite being a supervised method, Xlayer almost always performed worse than Lenses. It suggests that CNN feature maps contain richer information than class probabilities \cite{feat-distil}, and can be used to train another CNN more effectively.

\section{Conclusions}
We propose ``Feature Lenses'', an on-demand plug-and-play framework to counteract various image transformations towards invariant image representations. In particular, we present two lightweight lens designs to handle two common transformations: rotations and scalings, respectively. These designs can be easily extended to other transformations, such as changes of lighting conditions. Lenses are trained in a self-supervised fashion, by optimizing a novel ``Top-$K$ Activation Contrast Loss'' on feature maps. Feature Lenses significantly improved the accuracy of classifying distorted input images from ImageNet, MNIST-rot, and CIFAR-10, without retraining the original pipeline. We expect Feature Lenses to find broader applications, such as object detection and tracking, on which transformation-invariant representations have shown advantages \cite{rotinv-detection,snip}. 

\bibliographystyle{splncs04}
\bibliography{featlens}

\clearpage
\appendix

\section{Experimental Settings}
\subsection{Implementations of Methods}
\textbf{DataAug} The DataAug training code was based on the NVIDIA apex example training code\footnote{\url{https://github.com/NVIDIA/apex/tree/master/examples/imagenet}}. We added 90\degree, 180\degree~and 270\degree ~rotations, and 1/2, 1/3 scalings as additional augmentations. On ImageNet, the ResNet-101 model was trained for 90 epochs with FP16 on two RTX2080Ti GPUs for about one week.

\textbf{Xlayer}  The Xlayer model extends the torchvision ResNet library\footnote{\url{https://github.com/pytorch/vision/blob/master/torchvision/models/resnet.py}} by adding an extra Bottleneck or Basic block to layer4. During training, only this extra block is updated, and the lower layers of the pretrained ResNet are frozen. Two Xlayer instances were trained: one for handling the three types of rotations, and the other for handling the two scalings. On ImageNet, each instance was trained for 4 epochs with FP16 on one RTX2080Ti GPU for three hours.

\textbf{Lenses} We implemented Lenses in PyTorch. In order to insert Lenses into ResNet, we modified the torchvision ResNet code to ``intercept'' the features from the last Bottleneck/Basic block in layer4. During training, only the lenses are updated, and the pretrained ResNet is frozen.  Two sets of lenses were trained: one set of 3 rotation lenses, and one set of two scaling lenses. On ImageNet, each set was trained for 2 epochs with FP16 on one RTX2080Ti GPU for four hours.

\textbf{RP\_RF\_1} We obtained the source code of RP\_RF\_1 from the first author of \cite{backrot}. It consists of two custom layers written for Caffe. It was trained on the original MNIST and CIFAR-10, and tested on MNIST-rot and rotated CIFAR-10. 

\textbf{ORN} The first author of \cite{ORN} released their implementation online\footnote{\url{https://github.com/ZhouYanzhao/ORN/}}. It was trained on the original MNIST and CIFAR-10, and tested on MNIST-rot and rotated CIFAR-10.

Note that we trained RP\_RF\_1 and ORN on the original MNIST and CIFAR-10, instead of MNIST-rot and rotated CIFAR-10. The reason for  training on original images is that, if these rotation-invariant models receive rotated images as training images, then the test images are homogeneous to the training ones, which are not the scenarios they are designed for in the beginning. Only by training on original images and testing on rotated images, can the benefits of learning rotation-invariant features be revealed. In contrast, although Feature Lenses are trained on rotated images, the training is self-supervised without knowing their class labels (The training images even could be from a different dataset, such as Google Open Images\footnote{\url{https://storage.googleapis.com/openimages/web/index.html}}). So we believe they are evaluated on a level ground.

\subsection{Hyperparameters of Lenses}
The $K$ of the Top-$K$ Activation Contrast Loss is set to 6 for ImageNet, and 3 for MNIST-rot and rotated CIFAR-10, respectively, as the feature maps of ImageNet images are larger in size than those of MNIST-rot and rotated CIFAR-10. 

The overshoot discount factor $d_1$ in Eq.(5) in our manuscript is set to 0.2. This is to give less penalty to overshot features, as feature overshooting has less negative impact than undershooting to the downstream classifier.

\subsection{Training Settings}
On ImageNet, all models were trained with the SGD optimizer. On MNIST-rot and rotated CIFAR-10, DataAug, Xlayer and Lenses were trained with the Ranger optimizer\cite{ranger}.

Xlayer and Lenses have an initial learning rate of 0.01 on all datasets. On ImageNet, the learning rates of Xlayer and Lenses are decayed by 0.5 every 1/4 epochs. On MNIST-rot and rotated CIFAR-10, the learning rates are decayed by 0.5 every epoch. 

When training DataAug on ImageNet, each input batch is randomly transformed according to: 1) no transformation (but still applied with other augmentations) with a probability of 0.5,  2) 90/180/270\degree~ rotation, or 1/2 or 1/3 scaling, each with a probability of 0.1.

When training DataAug on MNIST-rot and rotated CIFAR-10, each input batch is randomly transformed according to: 1) no transformation (but still applied with other augmentations) with a probability of 0.4,  2) 90/180/270\degree~ rotation, each with a probability of 0.2.

In the above random transformation settings for DataAug, around half of the input images are untransformed. This is to avoid the trained models degrading significantly on original images. Nonetheless, the trained model degraded 2\% on original images under such a setting.

\subsection{Evaluation Criteria} 
The rotation accuracy on ImageNet (the ``Rot'' column in Table 3 in our submission) is defined as the average accuracy over images rotated by 90\degree, 180\degree~and 270\degree. 

The accuracy on MNIST-rot is defined as the average accuracy over images whose rotation angles are between [45\degree, 315\degree]. Images whose rotation angles are outside this range (i.e., within (-45\degree, 45\degree)) are very similar to the original MNIST images, on which all models achieved nearly 100\% accuracy. To highlight the performance on rotated images, these slightly-rotated images are not used to calculate the rotation accuracy. Therefore the accuracy of RP\_RF\_1 is slightly different from the accuracy reported in \cite{backrot}.

The accuracy on rotated CIFAR-10 is defined as the average accuracy over all images, including those slightly-rotated ones.

\subsection{A Brief Explanation of the Source Code}
\textbf{featlens.py} The library implementing Feature Lenses.

\noindent\textbf{train.py} The code to train and evaluate Lenses and Xlayer. Supports ImageNet, MNIST-rot and rotated CIFAR-10.

\noindent\textbf{train2.sh} The training script on two GPUs for Lenses or Xlayer.

\noindent\textbf{resnet.py} Modified code of ResNet, to accommodate with Lenses or Xlayer (controlled by initialization arguments).

\noindent\textbf{train-dataaug.py} Training and evaluation code of ResNet and DataAug. Supports ImageNet and MNIST.

\noindent\textbf{cifar10.py} Training and evaluation code of ResNet and DataAug on CIFAR-10.

\noindent\textbf{prepare\_data.py} The script to create ImageNet LMDB databases (slightly faster than loading from individual image files).

\noindent\textbf{lmdbloader.py} The library to load and augment ImageNet LMDB databases.

\noindent\textbf{ranger.py} The Ranger optimizer (\url{https://github.com/lessw2020/Ranger-Deep-Learning-Optimizer}).

\end{document}